\documentclass{article}

\usepackage[dblblindworkshop, final]{neurips_2026}

\workshoptitle{AI for Drug Discovery: Bridging the Translation Gap (AI4DD)}

\usepackage[utf8]{inputenc}
\usepackage[T1]{fontenc}
\usepackage{booktabs,amsmath,amsfonts,microtype,xcolor,hyperref,tikz}
\usepackage{graphicx}
\usepackage{dsfont}
\usetikzlibrary{arrows.meta,positioning}

\hypersetup{
    colorlinks=true,
    citecolor=blue!55!black,
    urlcolor=blue!55!black
}

\title{MoEDDI: Semantic Mixtures of Experts for Long-Tailed Drug--Drug Interaction Prediction}

\author{%
\textbf{Dinh Tan Nguyen}\textsuperscript{\normalfont 1,$\ast$},
\textbf{Quang-Hien Kha}\textsuperscript{\normalfont 2},
\textbf{Minh-Toan Dinh}\textsuperscript{\normalfont 3},
\textbf{Xuan-Huy Nguyen}\textsuperscript{\normalfont 4} \\
\textbf{Dac Phu Ho}\textsuperscript{\normalfont 5},
\textbf{Le Hoang Nguyen}\textsuperscript{\normalfont 6},
\textbf{Dang-Manh Ho}\textsuperscript{\normalfont 7},
\textbf{Sai Ho Ling}\textsuperscript{\normalfont 1,}\thanks{Corresponding authors.} \\[3pt]
\textsuperscript{1}University of Technology Sydney, Australia 
\textsuperscript{2}Taipei Medical University, Taiwan \\
\textsuperscript{3}National Central University, Taiwan 
\textsuperscript{4}Vietnam National University, Vietnam \\
\textsuperscript{5}University of Newcastle, Australia 
\textsuperscript{6}T\'el\'ecom Paris, France 
\textsuperscript{7}FPT University, Vietnam \\
\texttt{\{dinhtan.nguyen; steve.ling\}@uts.edu.au}
}

\begin{document}

\maketitle

\begin{abstract}
Descriptor-based drug--drug interaction (DDI) models typically apply a single global predictor despite heterogeneous physicochemical evidence and severely long-tailed labels. We introduce \textbf{MoEDDI (Mixture-of-Experts for Drug--Drug Interaction prediction)}, a chemically organized mixture-of-experts that routes overlapping descriptor families using low-dimensional statistics and combines their sparse top-$2$ mixture with a global T-DDI pathway through a residual logit correction. The residual parameterization can exactly preserve a supplied T-DDI checkpoint at initialization, although the evaluated models are trained from scratch. Across complete test partitions with 178, 86, and 92 classes, MoEDDI improves macro-F1 over reproduced seed-2026 T-DDI baselines from 0.8284 to 0.8485, 0.4729 to 0.6536, and 0.1807 to 0.4671. On the 3,780-feature corpus, the lowest-support group (89 classes with fewer than 50 test examples) shows the largest mean classwise gain (+3.30 F1 points). Calibration improves on one corpus but worsens on two. These single-seed results support semantic conditional specialization while motivating matched ablations and distribution-shift evaluation.
\end{abstract}

\section{Introduction}

Computational DDI prediction can prioritize candidate interactions for pharmacological review, but benchmark performance may not transfer to unseen drugs or chemical series. Long-tailed labels, drugs shared across data partitions, and poorly calibrated confidence further complicate translation. Evaluation should therefore consider minority-class performance, calibration, and distribution shift rather than aggregate accuracy alone.

Recent graph models, including SSI-DDI, GMPNN-CS, SA-DDI, and DSN-DDI, learn interaction-relevant molecular representations directly \citep{nyamabo2021ssiddi,nyamabo2022gmpnn,yang2022saddi,li2023dsn}. Descriptor-based models remain attractive because their inputs are deterministic, cacheable, and auditable. T-DDI reports 0.9434 accuracy and 0.8452 macro-F1 using an ensemble on 868,069 DrugBank-derived pairs spanning 178 interaction types \citep{kha2026tddi}. Its reported high-confidence macro-F1 of 0.8992 covers only 87.91\% of the test set and is therefore not a complete-test comparison.

In the released T-DDI data used here, no categorical columns are provided. Because TabTransformer attention operates over categorical embeddings \citep{huang2020tabtransformer}, prediction is effectively performed by the numerical MLP pathway. For $D=3{,}780$ and $C=178$, this pathway contains approximately 87.1M parameters and processes every pair through the same global predictor.

We ask whether chemically structured conditional specialization can complement this global pathway. MoEDDI partitions descriptors into overlapping semantic expert views, uses a low-bandwidth router for sparse top-$2$ mixing, and adds the resulting expert representation as a residual correction. Its residual head can be zero-initialized so that a supplied T-DDI checkpoint is reproduced exactly before adaptation.

Our contributions are: (i) overlapping chemically named descriptor experts; (ii) low-bandwidth top-$2$ routing using family-level statistics; (iii) a baseline-preserving residual parameterization; and (iv) complete-test evaluation across three corpora with macro-F1, long-tail, calibration, and router-utilization analyses. Results use one run per protocol on pair partitions that are neither drug- nor scaffold-disjoint; we therefore claim architectural promise rather than definitive state-of-the-art or translational superiority.

\section{Related Work}

\paragraph{DDI prediction.}
DeepDDI established large-scale multiclass DDI prediction from structural similarity profiles \citep{ryu2018deepddi}. More recent models increasingly operate on molecular graphs or learned substructures. SSI-DDI models interactions between drug substructures \citep{nyamabo2021ssiddi}, GMPNN-CS combines graph message passing with chemical-substructure information \citep{nyamabo2022gmpnn}, SA-DDI learns substructure-aware representations \citep{yang2022saddi}, and DSN-DDI explicitly studies generalization to unseen drugs \citep{li2023dsn}. These methods demonstrate the value of representation learning, but fixed molecular descriptors remain useful when deterministic preprocessing, caching, and feature-level auditability are desirable. MoEDDI focuses on this descriptor setting and asks whether known feature families can provide a useful inductive bias for conditional computation.

\paragraph{Mixtures of experts and tabular models.}
Sparse mixture-of-experts models conditionally allocate model capacity across inputs \citep{shazeer2017moe}, while later work highlights routing imbalance and instability as practical concerns \citep{fedus2022switch}. MoEDDI differs from conventional MoE designs in that expert identity is not learned from an arbitrary partition of the hidden space: each expert corresponds to a predefined descriptor family, and overlapping features may participate in more than one expert. The router is deliberately restricted to family-level summary statistics rather than the full descriptor vector. This separates the routing decision from high-dimensional prediction and provides a simple way to audit which expert families are selected for each pair. Unlike computationally sparse MoE systems, the present implementation evaluates all experts before applying top-$2$ weights; sparsity therefore induces specialization rather than inference-time FLOP savings.

\section{Methods}
\label{sec:method}

\begin{figure}[t]
\centering
\includegraphics[width=\linewidth]{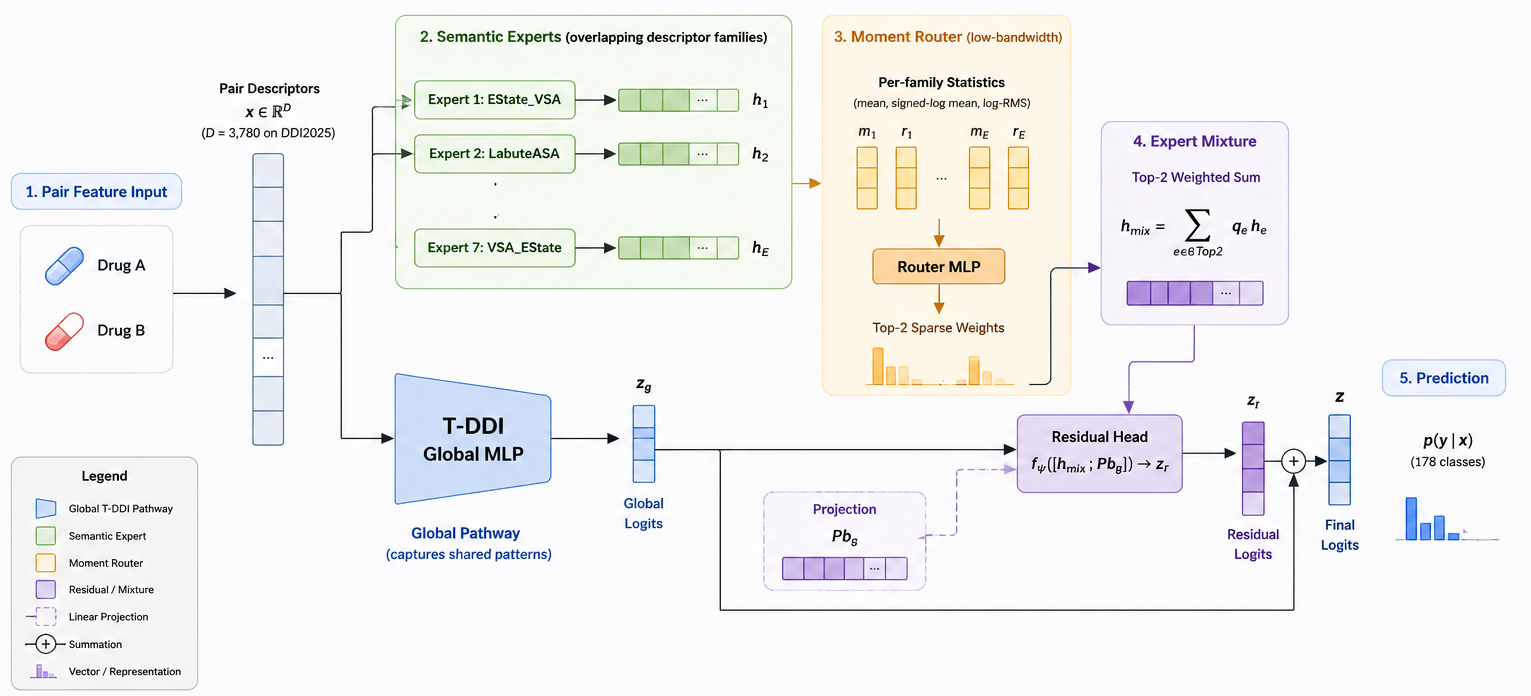}
\caption{MoEDDI on DDI2025. Semantic experts provide a pair-dependent residual correction to the global T-DDI pathway.}
\label{fig:architecture}
\end{figure}

\paragraph{Semantic expert views.}
Figure~\ref{fig:architecture} summarizes the architecture and its notation. Given pair features $\mathbf{x}\in\mathbb{R}^{D}$ and DDI label $y$, the model estimates $p(y\mid\mathbf{x})$. Identifiers, names, and SMILES are excluded. On DDI2025, descriptor names define seven overlapping expert sets $S_e$: EState\_VSA, LabuteASA, MR\_VSA, MTPSA, PEOE\_VSA, SlogP\_VSA, and VSA\_EState. Cross-family descriptors may belong to multiple experts, and labels do not define the partition. Each expert produces
\begin{equation}
\mathbf{h}_e=\phi_e(\mathbf{x}_{S_e})\in\mathbb{R}^{d_h},
\end{equation}
where $\phi_e$ is an independent LayerNorm--MLP; the main model uses $d_h=384$.

\paragraph{Design rationale.}
The architecture reflects three design choices. First, expert sets are allowed to overlap because several descriptor names encode more than one chemical family; forcing a disjoint partition would make expert membership depend on an arbitrary tie-breaking rule. Second, routing uses only two moments per family rather than the complete descriptor vector. The router therefore determines which representation should be emphasized from coarse family-level evidence, while the experts retain responsibility for high-dimensional feature extraction. Third, the global pathway is preserved on DDI2025 so that expert specialization acts as a correction rather than requiring the model to relearn an already strong global representation. This construction also permits exact recovery of a pretrained baseline at initialization when the residual output layer is zero-initialized.

\paragraph{Low-bandwidth top-$2$ routing.}
For each family, the router receives only signed-log mean and RMS summaries:
\begin{equation}
\bar{x}_e=\frac{1}{n_e}\sum_{j\in S_e}x_j,\qquad
m_e=\operatorname{sgn}(\bar{x}_e)\log(1+|\bar{x}_e|),\qquad
r_e=\log\left(1+\sqrt{\frac{1}{n_e}\sum_{j\in S_e}x_j^2}\right).
\end{equation}
A two-layer MLP maps $\mathbf{s}=[m_1,r_1,\ldots,m_E,r_E]$ to logits $\mathbf{a}$. Only the two largest logits are retained:
\begin{equation}
q_e=
\frac{
\exp(a_e)\mathds{1}[e\in\operatorname{Top2}(\mathbf{a})]
}{
\sum_j\exp(a_j)\mathds{1}[j\in\operatorname{Top2}(\mathbf{a})]
},
\qquad
\mathbf{h}_{\mathrm{mix}}=\sum_e q_e\mathbf{h}_e.
\label{eq:router}
\end{equation}
Dense pre-mask probabilities regularize router balance; sparse $q_e$ values are used for utilization analysis.

\paragraph{Baseline-preserving residual.}
Let $\mathbf{b}=g_{\theta}^{\mathrm{hid}}(\mathbf{x})$ and $\mathbf{z}_g=g_{\theta}^{\mathrm{cls}}(\mathbf{b})$ denote the global hidden representation and logits. MoEDDI predicts
\begin{equation}
\mathbf{z}_r=f_{\psi}([\mathbf{h}_{\mathrm{mix}};P\mathbf{b}]),
\qquad
\mathbf{z}=\mathbf{z}_g+\mathbf{z}_r.
\label{eq:residual}
\end{equation}
Zero-initializing the final layer of $f_\psi$ gives $\mathbf{z}_r=\mathbf{0}$, so a supplied T-DDI checkpoint is reproduced exactly at initialization. This is a parameterization property; the evaluated DDI2025 model is trained from scratch. The two lower-dimensional corpora instead use DrugA, DrugB, and Generalist views without the large residual backbone.

\paragraph{Training objective.}
Class imbalance is addressed with focal loss ($\gamma=1$) \citep{lin2017focal}. The DDI2025 objective supervises combined, residual, and global logits while regularizing router balance and logit magnitude:
\begin{equation}
\begin{split}
\mathcal{L}
={}&
\mathcal{L}_{\mathrm{focal}}(\mathbf{z},y)
+0.3\mathcal{L}_{\mathrm{focal}}(\mathbf{z}_r,y)
+0.3\mathcal{L}_{\mathrm{focal}}(\mathbf{z}_g,y)\\
&+10^{-2}\left(E\sum_e\bar{p}_e^2-1\right)
+10^{-3}\mathbb{E}_i
\left[
\log\sum_e\exp(a_{ie})
\right]^2,
\end{split}
\label{eq:loss}
\end{equation}
where $E$ is the number of experts, $\bar{p}_e$ is the batch mean of the dense (pre-mask) router probability $\operatorname{softmax}(\mathbf{a})_e$, and $a_{ie}$ is the router logit of expert $e$ for pair $i$. Checkpoints are selected by validation macro-F1 unless otherwise specified.

\section{Experimental Setup}

\paragraph{Datasets.}
We evaluate three fixed train/validation/test partitions (Table~\ref{tab:data}). DDI2025 is the DrugBank-derived corpus used by T-DDI \citep{wishart2018drugbank,kha2026tddi}; DeepDDI-86 uses a 100-dimensional pair representation for 86 interaction classes \citep{ryu2018deepddi}; FDA-DDI-92 contains 217 descriptors per drug, yielding 434 pair features. Preprocessing statistics are fit on training data only, checkpoint selection uses validation data, and every test pair is evaluated without confidence filtering.

\begin{table*}[t]
\caption{Corpora and fixed partitions.}
\label{tab:data}
\centering
\scriptsize
\setlength{\tabcolsep}{4pt}
\begin{tabular}{lrrrrrl}
\toprule
Corpus & Train & Validation & Test & Classes & Features & MoEDDI views \\
\midrule
DDI2025 & 520,841 & 173,614 & 173,614 & 178 & 3,780 & seven chemical families \\
DeepDDI-86 & 115,372 & 38,462 & 38,450 & 86 & 100 & DrugA, DrugB, Generalist \\
FDA-DDI-92 & 239,412 & 79,805 & 79,804 & 92 & 434 & DrugA, DrugB, Generalist \\
\bottomrule
\end{tabular}
\end{table*}

\paragraph{Baselines and evaluation.}
\textbf{T-DDI-MF} uses seed 2026 and validation macro-F1 selection; \textbf{T-DDI-Acc} is an available seed-42 accuracy-selected checkpoint. Both use the active numerical T-DDI MLP. Runs use mixed precision, batch size 256, learning rate $10^{-4}$, gradient clipping at 1, and focal loss. DDI2025 uses raw inputs for both models; on the two smaller corpora, MoEDDI uses training-set standardization whereas T-DDI retains raw inputs, so those comparisons are not strictly matched ablations. We report accuracy, macro-F1, weighted F1, top-$3$, top-$5$, and 15-bin ECE, with macro-F1 as the primary metric. DDI2025 classes are additionally grouped by test support ($<50$, 50--499, $\geq500$).

\section{Results and Discussion}
\label{sec:results}

\paragraph{Overall performance.}
Table~\ref{tab:results} reports complete held-out test performance. Under the seed-2026, macro-F1-selected protocols, MoEDDI improves macro-F1 from 0.8284 to 0.8485 on DDI2025, from 0.4729 to 0.6536 on DeepDDI-86, and from 0.1807 to 0.4671 on FDA-DDI-92. Accuracy increases by 0.94, 15.43, and 29.71 percentage points, respectively. Classwise win/tie/loss counts relative to T-DDI-MF are 94/44/40, 67/15/4, and 79/9/4.

\begin{table*}[t]
\caption{Complete held-out test performance. T-DDI-Acc denotes the T-DDI checkpoint selected by validation accuracy, whereas T-DDI-MF denotes the checkpoint selected by validation macro-F1. The two checkpoints also use different random seeds, as described in the text. Arrows indicate the preferred direction of each metric. Best within each corpus is bold.}
\label{tab:results}
\centering
\small
\setlength{\tabcolsep}{4pt}
\begin{tabular}{llrrrrrr}
\toprule
Corpus & Model & Accuracy $\uparrow$ & Macro-F1 $\uparrow$ & Weighted F1 $\uparrow$ & Top-3 $\uparrow$ & Top-5 $\uparrow$ & ECE $\downarrow$ \\
\midrule
DDI2025
& T-DDI-Acc & 0.9219 & 0.8004 & 0.9215 & 0.9934 & 0.9980 & \textbf{0.0044} \\
& T-DDI-MF & 0.9269 & 0.8284 & 0.9264 & 0.9930 & 0.9978 & 0.0213 \\
& MoEDDI & \textbf{0.9363} & \textbf{0.8485} & \textbf{0.9361}
& \textbf{0.9946} & \textbf{0.9982} & 0.0440 \\
\midrule
DeepDDI-86
& T-DDI-Acc & 0.5377 & 0.2540 & 0.5021 & 0.8474 & 0.9282 & \textbf{0.0534} \\
& T-DDI-MF & 0.6618 & 0.4729 & 0.6554 & 0.9185 & 0.9656 & 0.1122 \\
& MoEDDI & \textbf{0.8161} & \textbf{0.6536} & \textbf{0.8134}
& \textbf{0.9749} & \textbf{0.9912} & 0.0704 \\
\midrule
FDA-DDI-92
& T-DDI-Acc & 0.3198 & 0.0997 & 0.2198 & 0.5454 & 0.6705 & \textbf{0.0176} \\
& T-DDI-MF & 0.3371 & 0.1807 & 0.2495 & 0.5929 & 0.7076 & 0.0206 \\
& MoEDDI & \textbf{0.6342} & \textbf{0.4671} & \textbf{0.6092}
& \textbf{0.9153} & \textbf{0.9728} & 0.0475 \\
\bottomrule
\end{tabular}
\end{table*}

T-DDI-MF also exceeds the available T-DDI-Acc checkpoint in macro-F1, although seed and checkpoint criterion change simultaneously and therefore prevent a controlled selection ablation. On DDI2025, the published T-DDI ensemble reports 0.9434 accuracy and 0.8452 macro-F1 \citep{kha2026tddi}, compared with 0.9363 and 0.8485 for the single MoEDDI model. Because the training and ensemble protocols differ, this indicates competitiveness rather than controlled superiority. The published high-confidence macro-F1 of 0.8992 is likewise not directly comparable because it excludes 12.09\% of test pairs.

\paragraph{Long-tail performance.}
The DDI2025 improvement is concentrated in low-support classes. Among 89 classes with fewer than 50 test examples, mean per-class F1 increases from 0.728 to 0.761 (+3.30 points), compared with +0.70 points for classes with 50--499 examples and +0.80 points for classes with at least 500 examples. This pattern is consistent with conditional specialization being particularly useful in the label tail, although it does not establish a pharmacological mechanism.

\paragraph{Cross-corpus interpretation.}
The three corpora test related but not identical MoEDDI instantiations. DDI2025 retains the 87.1M-parameter T-DDI global pathway and learns an expert residual, providing the cleanest evidence that conditional specialization can complement a strong global predictor. The lower-dimensional corpora instead use DrugA, DrugB, and Generalist views without the large backbone. Their larger gains are encouraging, but MoEDDI also uses training-set standardization while T-DDI retains raw inputs; consequently, these margins cannot yet be attributed solely to routing. A standardized T-DDI control and matched schedules are required.

\paragraph{Routing and calibration.}
DDI2025 routing remains distributed rather than collapsing to one expert. Mean sparse router weights on the test set are 25.7\% MTPSA, 24.4\% MR\_VSA, 21.2\% LabuteASA, 19.1\% VSA\_EState, and 8.2\% EState\_VSA, whereas PEOE\_VSA (0.28\%) and SlogP\_VSA (1.07\%) are rarely selected. The smaller-corpus models favor the Generalist view (57.8\% on DeepDDI-86 and 59.8\% on FDA-DDI-92), and the DrugA/DrugB asymmetry (25.2\%/16.9\% and 27.9\%/12.3\%) motivates an explicit pair-swap test. These are behavioral summaries, not chemical attributions.

Calibration does not uniformly improve with discrimination. Relative to T-DDI-MF, MoEDDI reduces ECE on DeepDDI-86 (0.1122 to 0.0704) but increases it on DDI2025 (0.0213 to 0.0440) and FDA-DDI-92 (0.0206 to 0.0475). This reinforces that calibration should be evaluated separately from class-sensitive performance \citep{guo2017calibration}.

\paragraph{Limitations and translation.}
The current evidence is single-seed and not fully matched in preprocessing, optimization, or capacity. Stronger evaluation requires matched multi-seed runs, a capacity-matched residual MLP, dense/uniform and shuffled-family controls, warm-start versus scratch training, and unseen-drug, scaffold-disjoint, and temporal splits. Pair-split performance should not be interpreted as evidence of clinical safety: MoEDDI predicts curated interaction labels rather than patient-specific harm, dose effects, mechanism, or severity. Its intended role is prioritization for expert or experimental review, with future deployment requiring calibration, abstention, and monitoring under distribution shift.

\section{Conclusion}

MoEDDI introduces chemically structured conditional specialization into descriptor-based DDI prediction through overlapping semantic experts, low-bandwidth top-$2$ routing, and a residual parameterization capable of preserving a supplied T-DDI predictor at initialization. Across three complete test partitions, MoEDDI achieves higher macro-F1 than reproduced seed-2026 T-DDI baselines, including an increase from 0.8284 to 0.8485 on DDI2025, with the largest mean classwise gain in the lowest-support group. These results provide preliminary evidence that semantic mixtures can complement global descriptor models, while matched ablations and shift-aware evaluation remain necessary before stronger translational claims.

\section{Code and Model Availability}

Source code, trained models, and implementation resources for MoEDDI are publicly available.\\
\textbf{Source code:} \url{https://github.com/tommyngx/MoE_DDI}\\
\textbf{Models and weights:} \url{https://huggingface.co/TommyNgx/MoE_DDI/}

% ============================================================
% OPTIONAL CAMERA-READY ACKNOWLEDGMENTS
% Uncomment and complete if applicable.
%
% \begin{ack}
% We thank ...
%
% This work was supported by ...
% \end{ack}
% ============================================================

\clearpage

{\small
\setlength{\bibsep}{2pt}
\bibliographystyle{plainnat}
\bibliography{references}
}

\end{document}